\documentclass[11pt]{article}
\usepackage[final]{acl}
\usepackage{times}
\usepackage{latexsym}
\usepackage{amsmath}
\usepackage{amssymb}
\usepackage{booktabs}
\usepackage{array}
\usepackage{graphicx}
\usepackage{multirow}
\usepackage{algpseudocode}
\usepackage{flushend}
\usepackage{url}
\usepackage[T1]{fontenc}
\usepackage[utf8]{inputenc}
\usepackage{microtype}
\usepackage{inconsolata}

\title{The Menu Is an Execution Prior: State-Path Tool Menus for Online Agents}
\author{
  Bo Yan$^{1}$ \quad Weikai Lin$^{2}$ \quad Song Wang$^{1}$\thanks{ Corresponding author.}\\
  $^{1}$Department of Computer Science \& Institute of Artificial Intelligence, University of Central Florida \\
  $^{2}$Department of Computer Science, University of Rochester \\
  \texttt{\{bo949643,song.wang\}@ucf.edu} \quad \texttt{wlin33@ur.rochester.edu}
}
\hypersetup{
  pdftitle={The Menu Is an Execution Prior: State-Path Tool Menus for Online Agents},
  pdfauthor={Bo Yan, Weikai Lin, Song Wang}
}

\newcommand{\toolbench}{ToolBench}
\newcommand{\tooleval}{ToolEval}
\newcommand{\ourmethod}{State-Path Tool Menu}

\newcommand{\toolset}{\mathcal{T}}
\newcommand{\candidate}{\mathcal{C}}

\newcolumntype{L}[1]{>{\raggedright\arraybackslash}p{#1}}
\newcolumntype{C}[1]{>{\centering\arraybackslash}p{#1}}

\begin{document}
\maketitle

\begin{abstract}
Language models act through tools, yet practical agents face libraries containing thousands of interfaces.
We introduce the \emph{tool menu} as the short, ordered subset of available tools shown to an agent before execution.
The agent can call only tools in this menu.
Multi-step tasks require the final action and the prerequisite tools that create its inputs in a usable order.
Current constructors rank tools by request relevance, which can surface the final action while omitting or delaying less obvious producers.
We introduce the \emph{state path}, a pre-execution route from the observable request state to the desired outcome, and propose \ourmethod{} to learn it.
Our framework treats the menu as an \emph{execution prior} over these routes.
Its encoder represents which tools can run from the current state, how their outputs satisfy later inputs, and which orders recur in training paths.
A retriever covers an executable entry, the missing-input producers, and the final action.
A reranker then places producers before consumers.
On \toolbench{}, our menu raises online success from 0.737 to 0.898 and outperforms retrieval, reranking, generation, and routing baselines without changing the agent.
The State-Path menu also covers more complete chains with 32 tools than the official list covers with 128, and its success gain persists across executor families with different model capacities.
\end{abstract}

\begin{figure*}[t]
\centering
\includegraphics[width=\textwidth]{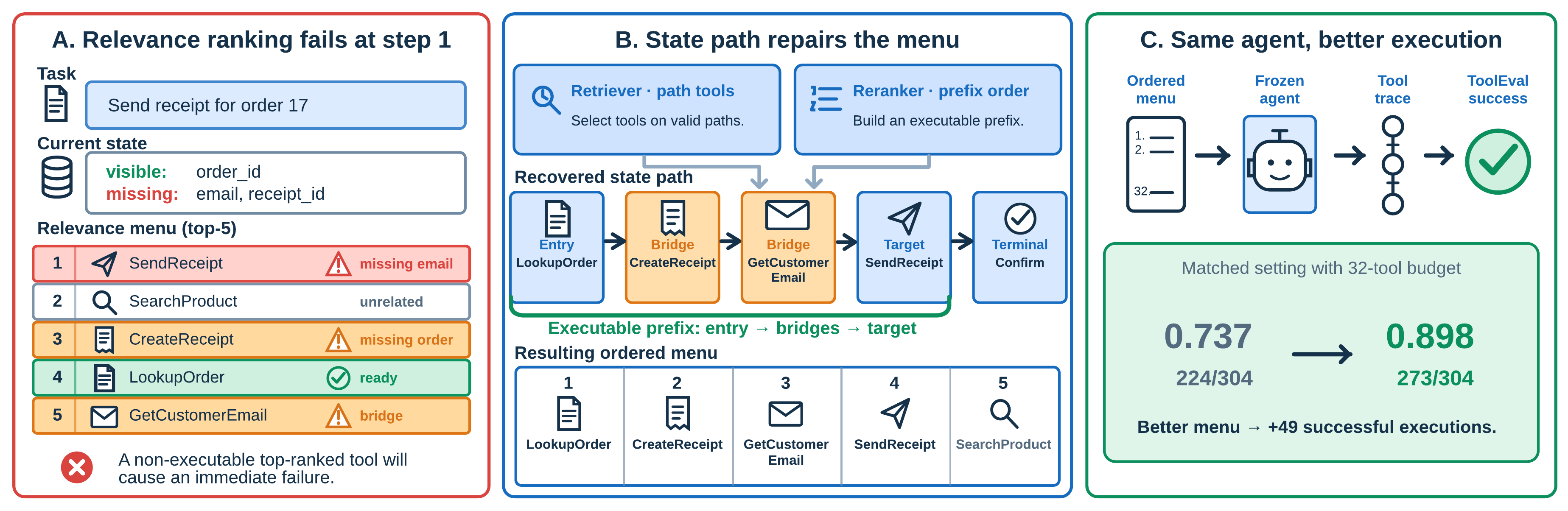}
\caption{\textbf{A relevance menu can expose the destination without a runnable route.}
For a receipt request, the answer-facing tool matches the text but depends on earlier tools that retrieve the order, create the receipt, and recover the recipient address.
\ourmethod{} uses the same menu budget to expose this route in executable order.
The online agent and call budget remain unchanged.}
\label{fig:teaser}
\end{figure*}

\section{Introduction}

Language models become agents by translating instructions into external actions \citep{Schick2023Toolformer,Yao2023ReAct}.
At each step, an agent selects a tool and uses its observation to choose the next action.
This loop becomes difficult as the action space grows.
\toolbench{}, for example, exposes more than 16{,}000 APIs \citep{Qin2024ToolLLM}.
Practical systems therefore retrieve a small candidate set before execution \citep{Patil2023Gorilla,Trivedi2024AppWorld}.
We introduce the term \emph{tool menu} for this request-specific, ordered interface.
It fixes the selected interfaces and their displayed order for the full execution while preserving the agent, its online execution loop, and the original call budget.

Most menu constructors rank tools by request relevance \citep{Shi2025ToolRet,Qu2024COLT,Zheng2024ToolRerank}.
Figure~\ref{fig:teaser} shows why relevance alone fails on multi-step tasks.
The left panel ranks \textsc{SendEmailReceipt} first before its receipt and address inputs exist.
The middle panel restores its prerequisite calls in executable order.
The right panel shows the resulting gain with the agent and 32-tool budget fixed.
The conflict is individual relevance against route completeness \citep{TRAJECTBench2026,Liang2026UniToolCall}.
This gap motivates a route-level construction objective for the bounded menu.

We therefore propose the \emph{state path}, a route through tool-induced states before execution.
It starts from visible fields, crosses tools that produce missing inputs, and ends at the requested outcome.
Individual relevance cannot express whether these calls form a complete route.
The state path makes this route explicit and turns the menu into an \emph{execution prior} before the first call.

To learn these paths, we propose \ourmethod{} with an encoder, a retriever, and a reranker.
The encoder learns relation-aware tool representations from the request state, tool schemas, and training paths.
The retriever uses these representations to select 32 tools with complementary path roles.
The reranker turns the selected set into producer-before-consumer order.
While acting on the resulting menu, the online agent keeps its original execution loop and call budget.

Our contributions are threefold:
\begin{itemize}
    \item We formulate the bounded, ordered tool menu as an execution prior and identify state-path completeness as the missing objective in relevance-based construction.
    \item We introduce \ourmethod{}, which jointly represents state compatibility, covers complementary path roles, and orders producers before consumers without changing the online agent or its call budget.
    \item We show that State-Path improves end-task success, complete-chain coverage, and executable entry across tool libraries, evaluators, and executor families. Paired outcomes and controlled variants connect these gains to route coverage and order within the same menu budget.
\end{itemize}

\begin{table*}[t]
\centering
\caption{\textbf{Comparison of the decisions made by tool-use method families.}
The columns show what each family selects, when selection occurs, which dependency information it uses, and whether execution updates that choice.}
\label{tab:capability-comparison}
\footnotesize
\setlength{\tabcolsep}{3.0pt}
\begin{tabular}{@{}L{0.22\textwidth}L{0.17\textwidth}C{0.14\textwidth}L{0.28\textwidth}C{0.13\textwidth}@{}}
\toprule
Method family & Selected object & Decision time & Dependency signal & Execution loop \\
\midrule
\toolbench{} official & ranked tools & before execution & query-to-tool text & unchanged \\
ToolRet / Tool-REX / COLT & tool set or ranking & before execution & text and query-to-tool structure & unchanged \\
ToolGen / Tool-Rank / SkillRouter & generated or ranked tools & before execution & identifiers, hierarchy, or text & unchanged \\
AutoTool / dynamic retrieval / ToolTree & next action or plan & during execution & plan state and observations & updated \\
\textbf{\ourmethod{}} & ordered state path & before execution & state, schema, and training paths & unchanged \\
\bottomrule
\end{tabular}
\end{table*}

\section{Related Work}

\paragraph{Tool-augmented language models and benchmarks.}
Toolformer and ReAct integrated tool calls into language-model generation and reasoning \citep{Schick2023Toolformer,Yao2023ReAct}.
Gorilla, HuggingGPT, ToolACE, and API-Bank extended tool use to larger and more varied interfaces \citep{Patil2023Gorilla,Shen2023HuggingGPT,Liu2025ToolACE,Li2023APIBank}.
\toolbench{} emphasizes library scale, AppWorld and $\tau$-bench require stateful execution, and UniToolCall and TRAJECT-Bench evaluate multi-step structure \citep{Qin2024ToolLLM,Trivedi2024AppWorld,Yao2024TauBench,Liang2026UniToolCall,TRAJECTBench2026}.
These settings make the tools visible before execution an object of study in their own right.

\paragraph{Tool retrieval, generation, and routing.}
ToolRet and ToolRerank improve query-to-tool matching through task-specific retrieval and hierarchy-aware reranking \citep{Shi2025ToolRet,Zheng2024ToolRerank}.
Tool-REX expands under-documented tool text, while SkillRouter studies full-text routing at registry scale \citep{Lu2026ToolREX,Zheng2026SkillRouter}.
ToolGen generates tool identifiers directly and removes the external retriever \citep{Wang2025ToolGen}.
All of these methods improve the identity or representation of tools selected for a request.

COLT is the closest method in its treatment of coverage.
It uses collaborative query-to-tool structure to recover a more complete set, although the selected tools remain unordered \citep{Qu2024COLT}.
State-Path requires each prerequisite to be reachable from observable state and to appear before its consumer in the displayed menu.

KELP reaches a related insight in knowledge-graph augmentation by scoring direct and indirect evidence paths against the input \citep{Liu2024KELP}.
KELP uses those paths to supply factual context.
State-Path uses state and schema direction to assemble a sequence of calls that can execute.
Table~\ref{tab:capability-comparison} compares the selected object, decision time, dependency signal, and execution loop across these method families under one pre-execution view.

\paragraph{Trajectory structure and adaptive tool planning.}
Dynamic Tool Dependency Retrieval conditions retrieval on an evolving function-calling plan \citep{Patel2026DynamicToolDependency}.
AutoTool uses historical transitions for tool selection, and ToolTree searches prospective branches with feedback from execution \citep{AutoTool2025,ToolTree2026}.
These planners update tool selection as the agent proceeds.

State-Path studies the route that can be exposed once before the first call while leaving the online loop unchanged.
Pre-execution construction provides the initial action space, and an online planner can revise that route after new observations.
The two decisions occur at different points in the agent loop and can be combined.
Choosing the menu once makes coverage and displayed order directly observable.
Appendix~\ref{sec:recent-planner-comparison} applies published dependency mechanisms to this pre-execution interface when their released design permits it.
Prior evidence that models are sensitive to buried information and option order also supports treating menu position as part of the interface \citep{Liu2023LostMiddle,Pezeshkpour2023OptionOrder}.

\section{Methodology}

We design \ourmethod{} to solve 2 failures created by a short menu.
Coverage fails when individually relevant tools omit a quiet bridge that creates a required input.
Order fails when the right tools are present but a consumer appears before a producer that makes it executable.
Our framework learns route membership and displayed order together as one pre-execution decision.

\begin{figure*}[t]
\centering
\includegraphics[height=0.35\textheight,keepaspectratio]{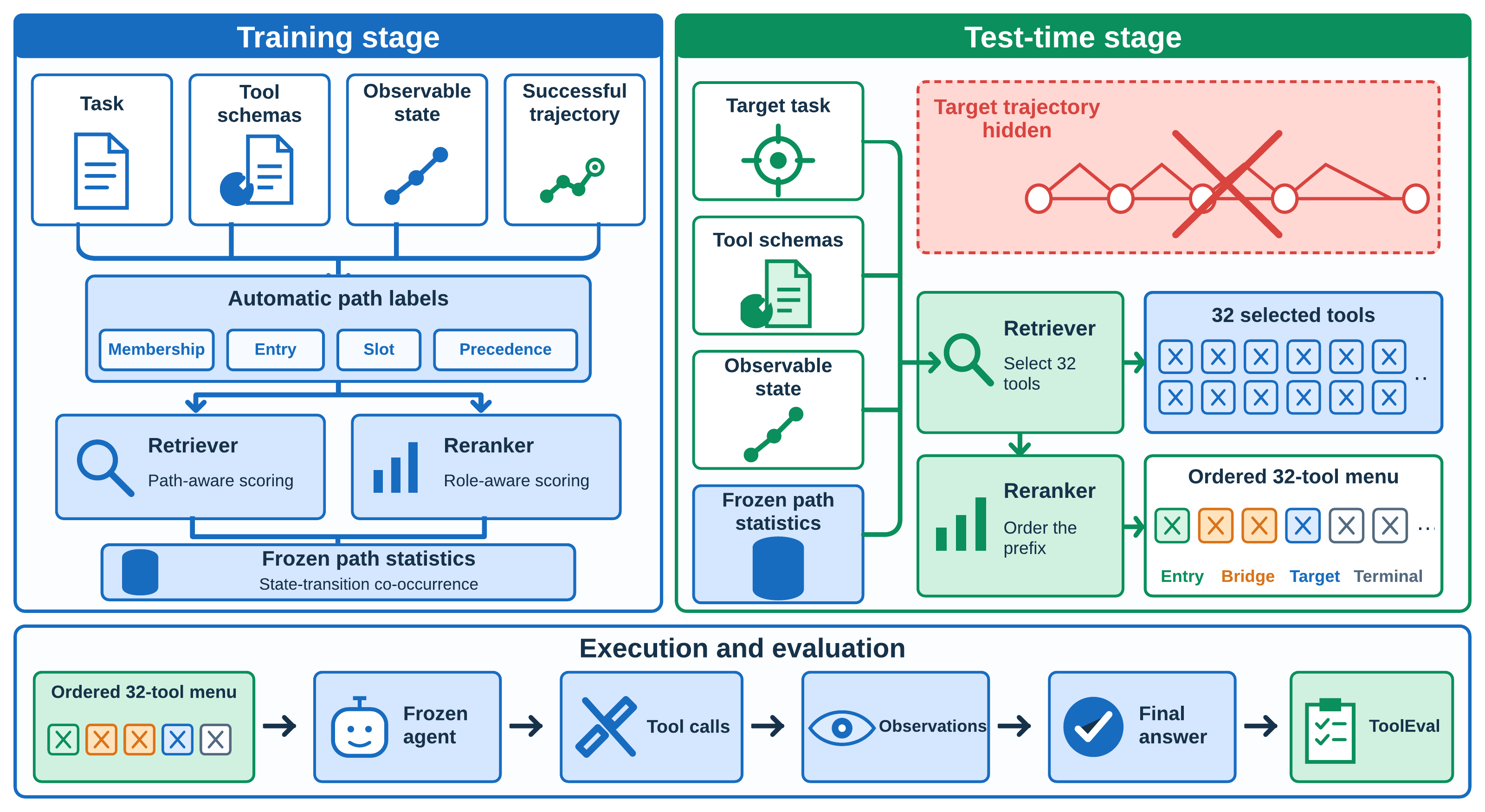}
\caption{\textbf{State-Path constructs a runnable route before execution.}
The retriever selects 32 tools that cover entry, bridge, target, and terminal roles.
The reranker places producers before consumers, and the online agent receives the completed menu once before its first call.}
\label{fig:method-overview}
\end{figure*}

Figure~\ref{fig:method-overview} traces this one-pass construction.
The upper left panel shows how training trajectories teach the constructor which tools enter a route and which calls precede others.
The upper right panel applies these learned relations to a new request whose target trajectory is hidden.
Our encoder first builds state- and relation-aware tool representations.
Our retriever uses them to cover complementary path roles.
Our reranker turns the selected set into an executable prefix.
The lower band shows that the online agent receives only the final menu while its execution loop remains unchanged.

The complete mapping is
\begin{equation}
\begin{aligned}
\candidate_0 &= \mathrm{Encode}(q,s_0(q),\toolset,\mathcal{M}),\\
\candidate_K &= \mathrm{Cover}_{R_\theta}(\candidate_0,K),\\
\pi_K &= \mathrm{Order}_{O_\psi}(\candidate_K,q,\mathcal{M}).
\end{aligned}
\label{eq:method-skeleton}
\end{equation}

Training trajectories provide supervision and path statistics.
The target trajectory remains hidden when the final menu is constructed.
We train the encoder and retriever with route membership, entry, and path-role targets.
We train the reranker with slot and pairwise precedence targets from the same trajectories.
This division gives the retriever responsibility for which tools enter the menu.
It gives the reranker responsibility for where those tools appear.
Appendix Table~\ref{tab:train-test-inputs} records the data available at each stage.
Appendix~\ref{sec:method-object-examples} follows the receipt example through all three components.

\subsection{Problem Formulation}
\label{sec:problem-formulation}

Let $\toolset=\{u_1,\ldots,u_N\}$ be a tool library.
Each tool $u_i$ carries document $d_i$, required inputs $I_i$, and produced outputs $O_i$.
A request $q$ exposes an initial state $s_0(q)$ containing the fields, objects, and constraints available before any call.
The constructor returns a menu $\pi(q)=(u_{\pi_1},\ldots,u_{\pi_K})$ with $K{=}32$.
The agent can invoke only tools in this ordered interface.
A tool is executable when its required inputs have already been observed or produced by earlier calls.

We define a \emph{state path} as a pre-execution hypothesis about an executable chain.
\begin{equation}
    \begin{aligned}
    s_0(q)&\rightarrow \textsc{entry}\rightarrow \textsc{bridge}^{*}\\
           &\rightarrow \textsc{target}\,[\rightarrow \textsc{terminal}].
    \end{aligned}
    \label{eq:state-path-definition}
\end{equation}
We assign each tool one of 4 roles along this route.
An entry tool can run from $s_0(q)$.
Bridge tools create inputs required later in the chain.
The target resolves the main request.
An optional terminal tool delivers or commits the result.
The target and terminal roles coincide when one call both resolves and commits the requested outcome.

For the receipt request, \textsc{LookupOrder} can use the visible order number.
\textsc{CreateReceipt} and \textsc{GetCustomerEmail} produce the fields required by \textsc{SendEmailReceipt}.
The sending tool serves as both the target and terminal action.
Its close lexical match becomes useful after the earlier calls provide its inputs.
This example separates textual relevance from executability.
The final action is easy to retrieve, while the bridge tools determine whether the route can run.

We write each successful training trajectory as $\tau=(g_1,\ldots,g_{T_\tau})$.
We learn membership, entry, role, and precedence from these trajectories and collect their path statistics in $\mathcal{M}$.
For a new request, the constructor must infer the hidden path from $q$, $s_0(q)$, tool schemas, and $\mathcal{M}$.
It then exposes the ordered menu to the agent.
Appendix~\ref{sec:full-method-details} collects the symbols and tensor definitions.

\subsection{State-Path Encoder}

To recover dependencies hidden by lexical similarity, we introduce 3 directional signals in the encoder.
They represent whether a tool can run now, whether it produces a field needed later, and which order recurs in training paths.
For visible fields $F_0(q)$ and tool inputs and outputs $I_i,O_i$, these signals are
\begin{align}
    \chi_i
    &= \frac{|I_i\cap F_0(q)|+\mathbb{I}[I_i=\varnothing]}
    {\max\{1,|I_i|\}},\\
    \kappa_{ij}
    &= \frac{|O_i\cap I_j|}{\max\{1,|I_j|\}}, \\
    \rho(i,j)
    &= \log\frac{c(i\prec j)+1}{c(j\prec i)+1}. \label{eq:main-transition}
\end{align}
Here $\chi_i$ measures whether $u_i$ can use the initial state.
Tools without required inputs have $\chi_i=1$ and can start from that state.
The schema score $\kappa_{ij}$ measures how much of $u_j$'s input can be produced by $u_i$.
The transition score $\rho(i,j)$ records their relative order in training trajectories.

These 3 signals distinguish tools that play different roles despite similar request relevance.
State compatibility identifies a runnable entry.
Schema flow identifies a producer-consumer link even when the tool names differ.
Path precedence strengthens links that recur across successful executions.

For the receipt example, state compatibility favors \textsc{LookupOrder} as the entry.
Schema flow connects the producer tools to \textsc{SendEmailReceipt}.
The transition score can recover a bridge whose name is absent from the request.

We feed candidate features and 4 path-summary tokens into a relation-aware Transformer.
We assign each ordered pair a relation type $r_{ij}$ for a forward dependency, reverse dependency, shared name or family, or no marked relation.
We inject this relation into attention as a learned bias.
\begin{equation}
    \alpha_{ij}\ \propto\ \exp\!\Big(\tfrac{\mathbf{q}_i^\top \mathbf{k}_j}{\sqrt{d}}+\beta(r_{ij})\Big).
    \label{eq:main-rel-attention}
\end{equation}
The resulting tool vectors carry local state fit and pairwise route structure.
The retriever consumes these vectors to choose route members.
The reranker also receives the directional features needed to order them.
Appendix~\ref{sec:full-method-details} gives the tensor shapes and training objectives.

\subsection{State-Path Retriever}

We design the retriever to preserve a complete route within 32 positions.
It begins with a broad frontier because a bridge tool can share few words with the request.
The membership score $M_i$ estimates whether $u_i$ belongs to the route.
The marginal coverage score $C_i$ rewards a new role or missing field.
The redundancy score $D_i$ penalizes overlap with tools already selected.
For the receipt request, these terms retain the customer-email bridge after the lookup tool enters the set.
These terms also prevent interchangeable terminal tools from displacing required bridges in the 32-tool menu.

An independent top-$K$ ranking can spend several positions on interchangeable terminal tools.
Our marginal decoder updates the uncovered roles and fields after every selection.
Once it selects an entry, producers for the remaining inputs receive more value than another tool with the same role.
The next slot maximizes the following score.
\begin{equation}
    s_i^{R}=M_i+\lambda C_i-\eta D_i.
    \label{eq:main-coverage-score}
\end{equation}
We select the highest-scoring candidate and update $C_i$ and $D_i$ for the next slot.
The decoder stops at 32 tools and passes this covered set directly to the reranker.
Appendix~\ref{sec:full-method-details} gives the full constrained objective and released decoder settings.

\subsection{State-Path Reranker}

We design the reranker because coverage alone can still expose a poor first action.
It receives exactly the 32 tools selected above and predicts their visible permutation.
The entry score $e_i$ favors a first tool that can run from $s_0$.
The slot score $\phi_{i,r}$ and precedence score $b_{ij}$ place producers before consumers.
Prefix state fit $P$ checks each proposed tool against the fields made available by the preceding prefix.
The reranker optimizes a leading executable prefix of length $L=8$.

We initialize the prefix state with the fields visible in the request.
After selecting a tool, we add its predicted outputs before scoring the next position.
This update lets an early producer make a later consumer executable.
It also prevents a static relevance order from placing the consumer first.
At position $r$, candidate $u_i$ receives the marginal score
\begin{equation}
    \begin{aligned}
    s_i^O(r)={}&w_m m_i+w_p\phi_{i,r}\\
    &+w_sP_i(s_{r-1})+\mathbb{I}[r=1]w_e e_i\\
    &+\frac{w_d}{\max\{1,r-1\}}\sum_{j<r}\left(b_{\pi_j i}-b_{i\pi_j}\right).
    \end{aligned}
    \label{eq:main-rerank-score}
\end{equation}
The final term averages net precedence support over the current prefix.
At the first position, the prefix is empty and the sum is zero.
The entry score acts only at the first position.
The precedence term then rewards producers that support the remaining consumers.
The decoder fills all 8 leading positions with this prefix-aware score.
The remaining 24 tools retain their retriever order as backups.
A bridge must enter the selected set before the reranker can place it ahead of its consumer.
The final output is the ordered menu passed unchanged to the online agent before execution begins.

\begin{table*}[!t]
\centering
\caption{\textbf{Main comparison of menu constructors.}
The \toolbench{} column reports online task success, while each remaining benchmark contributes its primary path measure.
Higher values are better.
Bold and underline mark the best and second-best values.}
\label{tab:main}
\footnotesize
\renewcommand{\arraystretch}{1.05}
\setlength{\tabcolsep}{10.0pt}
\begin{tabular}{lccccc}
\toprule
\multirow{2}{*}{Menu constructor}
& \toolbench{} & AppWorld & TRAJECT-Bench & UniToolCall & ToolHop \\
& Success & Next tool & Full chain & Ordered prefix & First action \\
\midrule
\toolbench{} official & \underline{0.737} & -- & -- & -- & -- \\
COLT
& 0.582 & \underline{0.190} & 0.230 & 0.215 & 0.238 \\
ToolRet
& 0.632 & 0.114 & 0.188 & 0.134 & 0.196 \\
Tool-REX family
& 0.661 & 0.180 & 0.350 & \underline{0.262} & \underline{0.597} \\
SkillRouter
& 0.707 & 0.123 & 0.167 & 0.169 & 0.467 \\
ToolGen
& 0.730 & 0.049 & \underline{0.708} & 0.215 & 0.265 \\
\midrule
\textbf{State-Path (ours)}
& \textbf{0.898}
& \textbf{0.465}
& \textbf{0.732}
& \textbf{0.635}
& \textbf{0.683} \\
\bottomrule
\end{tabular}
\end{table*}

\section{Experiments}

Can a better menu improve an agent without changing the agent itself?
We begin with this end-to-end question on \toolbench{}.
We then examine whether prerequisite coverage and executable order account for the improvement.
Further experiments change the evaluator, executor, and tool library to see how broadly the same path objective applies.
AppWorld, TRAJECT-Bench, UniToolCall, and ToolHop emphasize different demands on state, coverage, and order across these execution settings.

\subsection{Experimental Setup}
\label{sec:online-setting}

We first measure how much the menu alone changes online success on \toolbench{}.
Every method receives the same 304 released tasks and supplies one ordered menu of 32 tools before execution.
Qwen2.5-72B-AWQ uses that menu for the full run with deterministic decoding and an 8-call budget.
The prompt, execution wrapper, error handling, and official \tooleval{} evaluator are shared across all rows.

Training trajectories supervise the constructor, while development tasks select its constants.
The 304 final tasks contribute only the reported outcomes.
When a published method does not directly return a pre-execution menu, we apply its released selection rule and format the output as an ordered list of 32 tools.
Appendix~\ref{sec:evaluation-protocol} gives the full protocol, and Appendix~\ref{sec:baseline-adapters} describes each baseline adapter.

\subsection{Main Results}
\label{sec:main-results}

We compare State-Path with retrieval, reranking, generation, and routing methods in Table~\ref{tab:main}.
Every row supplies a 32-tool menu to the same online agent.
Our method solves 273 of the 304 \toolbench{} tasks, while the official menu solves 224.
The corresponding success rate increases from 0.737 to 0.898.
Among the 51 tasks on which the two menus disagree, State-Path wins 50 and loses 1 under the same evaluator across all paired tasks.

We then test different parts of a route on 4 additional benchmarks.
State-Path leads every primary measure in Table~\ref{tab:main}.
AppWorld and ToolHop reward a useful early action, UniToolCall rewards an ordered prefix, and TRAJECT-Bench requires a complete chain.
These measures trace a route at its entry, through its prefix, and at full-chain level.
Appendix Table~\ref{tab:full-cross-benchmark} contains the complete metric matrix for every method across all 5 benchmarks.

We use Table~\ref{tab:mechanism-attribution} to connect the success gain to route visibility.
Complete-chain coverage rises by 19.4 points, early-entry coverage by 11.8 points, and executable-prefix coverage by 3.8 points.
Our 32-tool menu also exceeds the official 128-tool list in complete-chain coverage.
The result establishes path composition as the decisive factor beyond menu width in this controlled comparison.

\begin{table}[!t]
\centering
\caption{\textbf{Route properties behind the \toolbench{} success gain.}
The first three rows compare 32-tool menus.
The last row compares the official 128-tool list with the 32-tool State-Path menu.}
\label{tab:mechanism-attribution}
\footnotesize
\setlength{\tabcolsep}{1.pt}
\begin{tabular}{@{}L{0.53\linewidth}C{0.14\linewidth}C{0.15\linewidth}C{0.13\linewidth}@{}}
\toprule
Diagnostic & Official & Ours& $\Delta$ \\
\midrule
Full chain in 32 tools & 0.510 & \textbf{0.704} & +0.194 \\
Entry tool in top 5 & 0.539 & \textbf{0.658} & +0.118 \\
Executable visible prefix & 0.296 & \textbf{0.334} & +0.038 \\
Official@128 vs. Ours@32 & 0.691 & \textbf{0.704} & +0.013 \\
\bottomrule
\end{tabular}
\end{table}

We partition the recovered failures in Table~\ref{tab:failure-conditioned}.
State-Path preserves 223 official successes and adds 50.
It loses 1 official success.
Of the 50 additions, 27 lacked a complete official chain and 11 lacked an early entry tool.
These 38 cases match the failures targeted by path coverage and prefix ordering.
The remaining 12 produce an accepted answer only with the State-Path menu.
These cases show that the executor's returned answer depends on the visible route it receives.

\begin{table}[!t]
\centering
\caption{\textbf{Breakdown of the 49-task net gain.}
Shares partition the 50 new successes by the failure recorded for the official menu.}
\label{tab:failure-conditioned}
\footnotesize
\setlength{\tabcolsep}{.pt}
\begin{tabular}{@{}L{0.7\linewidth}C{0.13\linewidth}C{0.13\linewidth}@{}}
\toprule
Paired outcome & Tasks & Share \\
\midrule
Official success preserved & 223 & -- \\
Official success lost & 1 & -- \\
New success w/ incomplete official chain & 27 & 54\% \\
New success w/ missing or late entry & 11 & 22\% \\
New success after answer rejection & 12 & 24\% \\
\bottomrule
\end{tabular}
\end{table}

\paragraph{Evaluator stability.}
\label{sec:metric-reliability}
\label{sec:judge-robustness}

We challenge the method ranking with 3 independent evaluators.
Eight menu constructors produce 400 traces on the first 50 executable tasks from the released hard split.
Three evaluators score these completed traces independently under their own decision criteria.

\begin{table}[!t]
\centering
\caption{\textbf{Comparison of method rankings across evaluators.}
Each evaluator scores the same completed traces, and State-Path ranks first in all three cases.}
\label{tab:judge-robustness}
\footnotesize
\setlength{\tabcolsep}{1.4pt}
\begin{tabular}{@{}L{0.32\linewidth}C{0.30\linewidth}C{0.20\linewidth}C{0.13\linewidth}@{}}
\toprule
Evaluator & Best external row & State-Path & Gap \\
\midrule
GPT-3.5-turbo-16k & 0.54 & \textbf{0.70} & +0.16 \\
GPT-4o-mini & 0.56 & \textbf{0.72} & +0.16 \\
GPT-5.5 & 0.30 & \textbf{0.34} & +0.04 \\
\bottomrule
\end{tabular}
\end{table}

All 3 evaluators place State-Path first in Table~\ref{tab:judge-robustness}.
GPT-5.5 assigns lower absolute scores to every method, yet State-Path remains 4 points ahead of the strongest external row.
GPT-4o-mini agrees with the released GPT-3.5-turbo-16k decisions on 91\% of the 400 outputs.
The 36 disagreements divide evenly between the 2 directions.
These results establish a stable leading position across evaluator thresholds and scoring scales.

\subsection{Ablation Study}

We ablate candidate coverage and tool ordering to test the contribution of each decision.
One variant keeps the 32 official candidates and changes only their order.
Another uses one fixed path template for every request.
Table~\ref{tab:ablation} compares these variants with the official menu and the complete method.

\begin{table}[!t]
\centering
\caption{\textbf{Ablation study of candidate coverage and tool ordering.}
Each row changes how tools enter the menu, how they are ordered, or both.
The agent, 32-tool budget, and evaluator remain unchanged.}
\label{tab:ablation}
\footnotesize
\setlength{\tabcolsep}{1.7pt}
\begin{tabular}{@{}L{0.34\linewidth}C{0.25\linewidth}C{0.22\linewidth}C{0.13\linewidth}@{}}
\toprule
Menu variant & Candidate coverage & Displayed order & Success \\
\midrule
Official menu & official retrieval & released & 0.737 \\
Official tools + path order & official retrieval & learned path order & 0.773 \\
Single-template path menu & fixed path template & template order & 0.773 \\
\textbf{\ourmethod{}} & learned path coverage & learned path order & \textbf{0.898} \\
\bottomrule
\end{tabular}
\end{table}

Changing only the order of the official candidates raises success from 0.737 to 0.773, showing that order helps when the required tools are already present.
The fixed-template menu also reaches 0.773 and confirms the value of recurring route shape.
The complete method reaches 0.898 after learning both route membership and request-specific order.
Its 12.5-point gain over either partial variant establishes the value of learning these 2 decisions together for online success.

\begin{figure}[!t]
\centering
\includegraphics[width=\columnwidth]{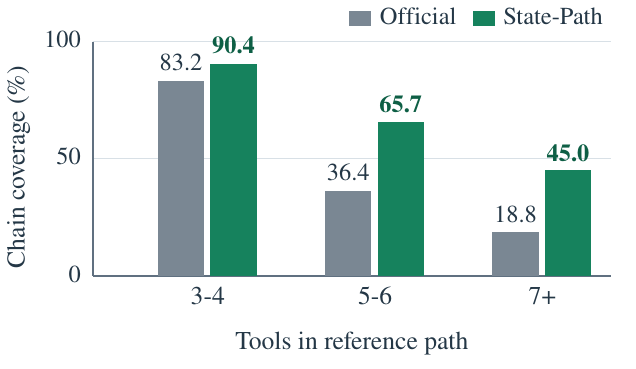}
\caption{\textbf{Complete-chain coverage by reference-path length.}
With 32 tools, the State-Path advantage grows from 7.2 points on 3--4-tool paths to 26.2 points on paths with at least 7 tools.}
\label{fig:path-length-coverage}
\end{figure}

We next test the prediction that path-aware construction matters most when a task contains more bridge tools.
Figure~\ref{fig:path-length-coverage} groups the 304 tasks by reference-path length under the same 32-tool menu width.
The coverage advantage grows from 7.2 points on 3--4-tool paths to 26.2 points on paths with at least 7 tools.
The fixed width isolates the growing competition among prerequisites.
This trend confirms that State-Path preserves quiet bridge tools as routes grow.

\subsection{Further Analysis}
\label{sec:evidence-availability}

\paragraph{Training-path evidence.}
State-Path combines tool schemas with recurring training paths.
To measure what the paths add, we rebuild their statistics with 0\%, 5\%, 10\%, 25\%, 50\%, and 100\% of the available trajectories.
Each nonzero share averages 5 samples.
Two corruption controls alter one step in 25\% or 50\% of the full paths.
Table~\ref{tab:path-memory-dose} shows how path quality affects coverage and recall.

\begin{table}[!t]
\centering
\caption{\textbf{Effect of training-path amount and corruption on route recovery.}
Subset rows average five samples, and every row excludes target-task trajectories.}
\label{tab:path-memory-dose}
\footnotesize
\setlength{\tabcolsep}{2.0pt}
\begin{tabular}{@{}L{0.37\linewidth}C{0.16\linewidth}C{0.19\linewidth}C{0.19\linewidth}@{}}
\toprule
Training evidence & Share & Chain@32 & Tool recall \\
\midrule
No path memory & 0\% & 0.510 & 0.747 \\
Clean subset & 5\% & 0.518 & 0.742 \\
Clean subset & 10\% & 0.547 & 0.756 \\
Clean subset & 25\% & 0.589 & 0.773 \\
Clean subset & 50\% & 0.636 & 0.794 \\
Clean full memory & 100\% & 0.704 & 0.827 \\
25\% corrupted & 100\% & 0.701 & 0.825 \\
50\% corrupted & 100\% & 0.694 & 0.823 \\
\bottomrule
\end{tabular}
\end{table}

Complete-chain coverage rises steadily as clean training paths are added in Table~\ref{tab:path-memory-dose}.
The schema-only constructor already recovers 0.510 of the complete chains when no path memory is available.
The full trajectory set raises this value to 0.704, a 19.4-point gain over schemas alone.
Corrupting half of the full set reduces coverage by only 1.0 point.
The increasing data curve establishes the contribution of recurring relations.
The corruption control shows that repeated evidence preserves this contribution under frequent local path errors.

\paragraph{Construction cost.}

We measure the quality and latency trade-off introduced by menu construction.
All measurements use batch size one on a single B200.
Table~\ref{tab:construction-cost} separates retrieval, path coverage, feature construction, constrained decoding, and the 2 neural passes.

\begin{table}[!t]
\centering
\caption{\textbf{Breakdown of per-request menu construction time on one B200.}
Constructor rows cover 284 valid frontiers.
Neural rows use 1,000 timed forwards after 20 warm-up runs with precomputed features.}
\label{tab:construction-cost}
\footnotesize
\setlength{\tabcolsep}{2.4pt}
\begin{tabular}{@{}L{0.46\linewidth}C{0.16\linewidth}C{0.16\linewidth}C{0.15\linewidth}@{}}
\toprule
Component & Median ms & P95 ms & Params \\
\midrule
Live semantic retrieval & 13.63 & 16.85 & -- \\
Path coverage & 139.34 & 193.99 & -- \\
Features and order decode & 266.11 & 368.39 & -- \\
Cached-frontier constructor & 406.41 & 552.85 & -- \\
Neural retriever & 2.61 & 2.88 & 0.416M \\
Neural reranker & 3.35 & 4.56 & 0.449M \\
\bottomrule
\end{tabular}
\end{table}

The complete constructor remains below 0.6 seconds at the 95th percentile.
The two neural passes together take about 6 ms.
Most of the measured time comes from explicit path-feature construction and constrained decoding.
The menu is built once, so this cost is shared by all subsequent calls.
The learned models therefore contribute little to the latency, while the explicit route constraints account for most of the one-time cost.

\paragraph{Tool documentation.}
We test which documentation fields supply the relations used by the schema frontier.
Table~\ref{tab:documentation-signal} compares names alone, names with structured input and output fields, and full descriptions on the same released library.
Adding input and output fields improves dependency-edge accuracy by 4.8 points and path-replay success by 5.0 points on the same released library.

Full descriptions match the structured-field condition on both measures.
This match identifies the structured fields as the useful dependency signal.
Names describe what a tool does, while input and output fields reveal which calls can connect.
The result explains why schemas provide a viable path frontier before trajectory evidence accumulates.

\begin{table}[!t]
\centering
\caption{\textbf{Effect of tool-documentation fields on dependency prediction and replay.}
All three variants use the same released tool library.}
\label{tab:documentation-signal}
\footnotesize
\setlength{\tabcolsep}{3.0pt}
\begin{tabular}{@{}L{0.48\linewidth}C{0.24\linewidth}C{0.21\linewidth}@{}}
\toprule
Documentation & Edge accuracy & Replay success \\
\midrule
Tool names & 0.628 & 0.105 \\
Names + input/output fields & \textbf{0.676} & \textbf{0.155} \\
Full descriptions & \textbf{0.676} & \textbf{0.155} \\
\bottomrule
\end{tabular}
\end{table}

\paragraph{Tool-library generalization.}
\label{sec:transfer}
\label{sec:cross-dataset}

We first apply the same architecture and objective within 4 benchmark splits.
Table~\ref{tab:main} shows that State-Path leads every primary measure.
The largest margins occur when a metric requires an executable entry or an ordered prefix within a bounded menu.
This pattern extends the route objective beyond the \toolbench{} library and its end-task evaluator.

We then test whether relations learned in one tool library remain useful in another.
We select relation weights on ToolBench and apply them unchanged to 2,000 hard TRAJECT-Bench tasks.
No target-library trajectories or target labels are used for model selection.
Table~\ref{tab:frozen-cross-library} compares this frozen transfer with semantic retrieval on the target library.

\begin{center}
\begin{minipage}{\columnwidth}
\centering
\captionof{table}{\textbf{Cross-library transfer without target-library training.}
Relation weights are selected on \toolbench{} and then applied to TRAJECT-Bench without target trajectories or labels.}
\label{tab:frozen-cross-library}
\footnotesize
\setlength{\tabcolsep}{2.5pt}
\begin{tabular}{@{}L{0.39\linewidth}C{0.18\linewidth}C{0.18\linewidth}C{0.17\linewidth}@{}}
\toprule
Constructor & Chain@32 & Tool recall & First tool \\
\midrule
Semantic retrieval & 0.247 & 0.625 & 0.703 \\
Frozen relation prior & \textbf{0.254} & \textbf{0.636} & \textbf{0.725} \\
\bottomrule
\end{tabular}
\end{minipage}
\end{center}

All three measures improve, with the largest change appearing in first-tool coverage at 2.2 points.
The largest gain at the first tool establishes a shared notion of executable entry across libraries.
The smaller gains in full-chain coverage and tool recall identify where local trajectories add library-specific route evidence.

\begin{figure}[!t]
\centering
\includegraphics[width=\columnwidth]{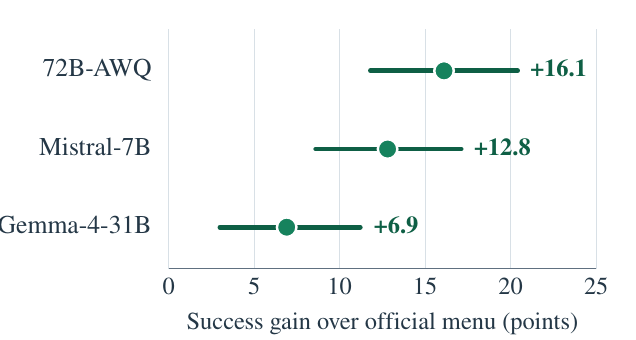}
\caption{\textbf{State-Path improves success across executors.}
Points show paired gains over the official menu for three executor families.}
\label{fig:executor-robustness}
\end{figure}

\paragraph{Executor families.}
We finally ask whether the menu gain depends on the executor family.
Three executors receive the same ToolBench tasks and the same pair of official and State-Path menus.
Figure~\ref{fig:executor-robustness} reports paired success gains of 16.1, 12.8, and 6.9 points.
All 3 gains are positive.
The result shows that a visible route benefits different model capacities and planning styles.

\section{Conclusion}

The short tool menu determines whether an online agent can see a usable route before the first call.
\ourmethod{} constructs that route by encoding observable state and tool dependencies, covering the required path roles, and placing producers before consumers.
On \toolbench{}, changing only the menu raises success from 0.737 to 0.898 and recovers 49 additional tasks.
Chain, entry, and prefix measurements connect these recoveries to the proposed mechanism.
Independent evaluators with different decision criteria confirm the same advantage across tool libraries and executor families.

\section{Limitations}

State-Path fixes one menu before execution, leaving observation-conditioned updates open.
The constructor relies on informative input and output fields and recurring path relations.
Its gradual degradation with sparse or corrupted histories motivates learning relation evidence in new or lightly documented libraries.
Future work can add online updates while retaining initial route coverage.

\section{Ethics Statement}

The experiments use public research benchmarks, licensed models, and synthetic API tasks that contain no personal data.
Improving route visibility can also make harmful tool sequences easier to execute when a request or tool library is unsafe.
Practical deployment therefore requires permission checks, sandboxed execution, and human review for consequential actions in real deployments.

\bibliography{references}

\appendix

\section*{Appendix}

\section{Evaluation Protocol}
\label{sec:evaluation-protocol}

The main comparison contains 304 released \toolbench{} tasks after applying the official split loader and executable-tool filter.
Every row presents a different ordered menu to the same execution agent; the execution wrapper, call budget, and evaluator stay fixed throughout the comparison.

\subsection{Training and Test Inputs}

Table~\ref{tab:train-test-inputs} records when each information source is available.
Successful trajectories provide supervision on training tasks.
For a final task, the constructor receives the request, observable state, tool schemas, and statistics derived from the training split.
The constructor selects each final menu without the target trajectory or any evaluator decision.

\begin{center}
\centering
\captionof{table}{\textbf{Information available during training and final menu construction.}
The target trajectory and evaluator feedback are hidden when a final menu is built.}
\label{tab:train-test-inputs}
\footnotesize
\setlength{\tabcolsep}{3pt}
\begin{tabular}{@{}L{0.24\linewidth}L{0.32\linewidth}L{0.32\linewidth}@{}}
\toprule
Item & Training stage & Test-time stage \\
\midrule
Task text & training request $q$ & target request $q$ \\
Tool library and schemas & visible & visible \\
Observable state $s_0(q)$ & extracted from request and schemas & extracted from request and schemas \\
Successful trajectory & visible for training tasks & absent for target task \\
Path statistics $\mathcal{M}$ & built from training trajectories & reused after training \\
Supervision & membership, entry, slot, precedence & none \\
Model output & trained retriever and reranker & ordered 32-tool menu $\pi_K$ \\
\bottomrule
\end{tabular}
\end{center}

\subsection{Online Execution Setting}

The online setting holds the following items constant across menu constructors.
\begin{itemize}
    \item The task set contains 304 released tasks in groups of 107, 134, and 63.
    \item The interface contains one ordered 32-tool menu built before execution.
    \item Each run allows 8 tool calls. Invalid and valid calls consume the same budget.
    \item The execution agent is Qwen2.5-72B-AWQ with deterministic decoding.
    \item Every run uses the same timeout, error handling, empty-response handling, and official binary \tooleval{} success measure.
    \item The random seed is 20260525.
\end{itemize}
The evaluator does not receive the identity of the menu constructor.

\subsection{Agent Prompt, Tool Failures, and Evaluator Input}

The agent receives the user task and the ordered menu.
Each menu entry contains a tool name, description, input schema, and argument format.
The same menu-neutral instruction is used in every run.

\begin{quote}
\footnotesize
You are a tool-using assistant.
Use only tools in the provided menu.
At each step, either call one tool with a valid JSON argument or produce the final answer.
Use earlier observations when deciding the next action.
Stop when the user request is satisfied or the step budget is exhausted.
\end{quote}

Missing tools, invalid arguments, tool exceptions, and empty responses are converted into standardized observations under the same 8-step budget.
The \tooleval{} evaluator receives the task, trace, observations, and final answer.
It returns one binary success label.

\subsection{Baseline Implementations}
\label{sec:baseline-adapters}

Every external baseline produces one ordered 32-tool menu before execution.
Retrieval baselines retain their native scores and are truncated to 32 tools.
Organizer baselines reorder a common candidate budget without using tool calls, execution observations, or evaluator feedback.

We retain released hyperparameters when they are available.
When a paper does not define a pre-execution menu, development tasks determine how its released selection rule produces 32 tools.
The chosen implementation is then used unchanged for final evaluation.
State-Path decoder constants follow the same development-only procedure.

The adapters fall into three groups.
\begin{itemize}
    \item The official retriever, COLT-Contriever, ToolRet-E5, and Tool-REX Embed construct their own candidate menus.
    \item ToolGen and SkillRouter reorder a shared candidate pool with constants selected on development tasks.
    \item Tool-Rank uses its released ranking weights on the same shared candidate interface.
\end{itemize}

The main table includes methods that construct their own candidates and methods that organize a shared candidate pool.
The named rows follow the \toolbench{} official retriever \citep{Qin2024ToolLLM}, COLT \citep{Qu2024COLT}, ToolRet \citep{Shi2025ToolRet}, Tool-REX and Tool-Rank \citep{Lu2026ToolREX}, ToolGen \citep{Wang2025ToolGen}, and SkillRouter \citep{Zheng2026SkillRouter}.

\subsection{Data Splits and Model Selection}

Training trajectories and tool schemas provide the path statistics $\mathcal{M}$.
Development tasks select the loss weights, decoder constants, hard-negative policy, and baseline-adapter constants.
Reranker training menus are generated out of fold by retrievers fitted without the corresponding training rows.
Development menus use the retriever selected on the training split.

The final 304 tasks are used for success reporting and diagnostics only.
Their reference chains and evaluator feedback do not update the constructor or its thresholds.
The public benchmarks, text encoder, and execution models are used under their research licenses.
The synthetic API tasks draw on no real-user records and contain no personal data.

\section{Worked Examples for Method Objects}
\label{sec:method-object-examples}

The receipt request illustrates each method object before the tensor definitions.
Its initial state contains only $F_0(q)=\{\texttt{order}\}$.
The menu must connect this visible field to the final sending action.
Table~\ref{tab:worked-example} maps the visible state, path roles, and required order in this example.

\begin{center}
\centering
\captionof{table}{\textbf{State-path objects in the running receipt example.}
Training trajectories supervise these roles, and the constructor predicts them for a new task.}
\label{tab:worked-example}
\footnotesize
\setlength{\tabcolsep}{3pt}
\begin{tabular}{@{}L{0.24\linewidth}L{0.66\linewidth}@{}}
\toprule
Object & Receipt example \\
\midrule
Visible state & \texttt{order} is present before execution \\
Entry & \textsc{LookupOrder} can run from \texttt{order} \\
Bridge & \textsc{CreateReceipt}, \textsc{GetCustomerEmail} create missing fields \\
Target / terminal & \textsc{SendEmailReceipt} resolves and delivers the request \\
Order & producers precede the sending tool \\
\bottomrule
\end{tabular}
\end{center}

Training paths in $\mathcal{M}$ often place \textsc{LookupOrder} before the two bridge tools and \textsc{SendEmailReceipt} after them.
State fit favors the lookup tool as the entry.
Schema flow links the producer tools to the sending tool.
Precedence keeps the sending action behind the calls that create its two required fields.

The resulting prefix is
\begin{center}
\footnotesize
\textsc{LookupOrder} $\rightarrow$ \{\textsc{CreateReceipt}, \textsc{GetCustomerEmail}\} $\rightarrow$ \textsc{SendEmailReceipt}.
\end{center}
In this example, state fit identifies the starting action.
Schema flow recovers the missing producers, and path precedence organizes the resulting menu.

\section{State-Path Implementation Details}
\label{sec:full-method-details}

Each task defines a relation context over its candidate tools.
A relation-aware Transformer predicts route coverage and executable order from that context.
The decoder converts these predictions into the ordered menu shown to the online agent.

\subsection{Notation}

Table~\ref{tab:notation} defines the symbols used in the equations and implementation details below.

\begin{center}
\centering
\captionof{table}{\textbf{Notation used in the implementation details.}}
\label{tab:notation}
\footnotesize
\setlength{\tabcolsep}{4pt}
\begin{tabular}{@{}L{0.26\linewidth}L{0.64\linewidth}@{}}
\toprule
Symbol & Meaning \\
\midrule
$q,\toolset$ & user task and tool library \\
$u_i,d_i$ & tool and document \\
$K,L,\pi(q)$ & menu budget, executable-prefix length, and ordered menu \\
$s_0(q)$ & observable state before execution \\
$F_0,I_i,O_i$ & visible fields, inputs, outputs \\
$\tau$ & training tool trajectory \\
$\mathcal{M}$ & frozen training path statistics \\
$U_q,\candidate_K$ & frontier and selected 32-tool set \\
$C_q^R,C_q^O$ & retrieval and ordering contexts \\
$r_{ij}^R,a_{ij}^O$ & relation type and pair feature \\
$m_i,e_i,\phi_{i,r},b_{ij}$ & membership, entry, slot, precedence scores \\
\bottomrule
\end{tabular}
\end{center}

\subsection{Feature Construction and Neural Hyperparameters}
\label{sec:feature-hyperparams}

All neural features are computed before execution.
The candidate features include semantic anchor scores, reciprocal ranks, state compatibility, missing-field counts, schema flow, training-path transition priors, and tool-family indicators.
We retain the first 64 dimensions of each frozen BGE representation and apply $\ell_2$ normalization for both queries and tools.
Queries use the BGE retrieval instruction, while tool documents are encoded directly as document-side inputs.

The target task's reference chain, evaluator decision, post-execution observations, and failure tag are excluded from these features.
Table~\ref{tab:neural-hyperparams} lists the model dimensions and optimization settings.

\begin{table}[!t]
\centering
\caption{\textbf{Released architecture and optimization settings.}
Development tasks select the tunable weights.
The dimensions define the released architecture.}
\label{tab:neural-hyperparams}
\footnotesize
\setlength{\tabcolsep}{4pt}
\begin{tabular}{@{}L{0.34\linewidth}L{0.58\linewidth}@{}}
\toprule
Parameter & Released value \\
\midrule
Text encoder & BGE-large-en-v1.5, frozen \\
Prepool / final menu / online steps & 128 / 32 / 8 \\
Negative sampling & hard, in-frontier, boundary \\
Feature dimensions & tool 64, query 64, retriever node 12, reranker node 14, pair 16 \\
Encoder & hidden 128, heads 4, layers 2 \\
Path tokens / slots / relation types & 4 / 8 / 5 \\
Prefix length / field capacity & 8 / 32 \\
Retriever anchors / neighbors / coverage / diversity & 64 / 4 / 0.45 / 0.12 \\
Training & 2 and 4 epochs, batch size 8 \\
Optimizer & AdamW, weight decay 0.01, dropout 0.08, clip 1.0 \\
Learning rates & retriever $8{\times}10^{-4}$, reranker $10^{-4}$ \\
Training seeds & retriever 20260525, reranker 17 \\
Reranker score: entry / member / position / state / direction & 0.55 / 0.35 / 0.65 / 0.60 / 0.90 \\
Reranker loss: entry / slot / precedence / member / path & 0.55 / 0.55 / 0.75 / 0.30 / 1.15 \\
\bottomrule
\end{tabular}
\end{table}

\subsection{Retriever and Reranker Equations}

Before execution, the observable state contains the fields visible in the request.
\begin{equation}
    s_0(q)=F_0(q),
    \label{eq:observable-state}
\end{equation}
where $F_0(q)$ is the set of visible fields in the request.

Pre-execution compatibility measures how much of a tool's input is already visible.
\begin{equation}
    \begin{aligned}
    \chi_i(s_0)&=
    \frac{|I_i\cap F_0(q)|+\mathbb{I}[I_i=\varnothing]}
    {\max\{1,|I_i|\}},\\
    \mathrm{miss}_i&=1-\chi_i(s_0).
    \end{aligned}
    \label{eq:state-compatibility}
\end{equation}

Training trajectories also provide a pairwise precedence label.
\begin{equation}
    y_{ij}^{prec}=\mathbb{I}[\mathrm{pos}_{\tau}(u_i)<\mathrm{pos}_{\tau}(u_j)].
    \label{eq:prec-label}
\end{equation}

\subsection{State-Path Retriever}

The retriever begins with a broad candidate frontier.
Let $f(\cdot)$ encode the task and each tool document.
\begin{equation}
    \mathrm{sim}_i = f(q)^\top f(d_i).
\end{equation}

High-scoring anchors $A_q$ are expanded with neighbors observed in training paths.
\begin{equation}
    U_q=A_q \cup \bigcup_{u_i\in A_q}\mathrm{Nbr}_{\mathcal{M}}(u_i)\cup H_q,
    \label{eq:frontier}
\end{equation}
where $H_q$ is a high-recall fallback.
In the released decoder, $H_q$ contains the remaining highest-scoring semantic candidates until the frontier reaches its released limit of 128 tools.

The expanded tools and learned path-summary tokens form the retriever relation context.
\begin{equation}
    C_q^R=(V_q^R,R_q^R),\quad
    V_q^R=\{p_1,\ldots,p_P\}\cup U_q,
    \label{eq:retriever-relation-context}
\end{equation}
where $p_1,\ldots,p_P$ are learned tokens that summarize recurring training paths.

For each tool pair, we compute a directional precedence score from the training paths
\begin{equation}
    \rho(i,j)=\log\frac{c(i \prec j)+1}{c(j \prec i)+1},
    \label{eq:transition-prior}
\end{equation}
and the relation context assigns one discrete type to each ordered tool pair in the retriever graph.
\begin{equation}
    r_{ij}^R\in\{0,1,2,3,4\}.
\end{equation}
The five values represent a neutral relation, a forward dependency, a reverse dependency, a shared name, and a shared family.
A training-path transition or schema flow establishes a directional dependency between the corresponding tools.

Each candidate receives the following compact feature vector.
\begin{equation}
    x_i^R =
    [\mathrm{sim}_i,\; \mathrm{rrank}_i,\; \mathrm{hist}_i,\; a_i^{in},\; a_i^{out},\; z_i].
\end{equation}
The value $\mathrm{hist}_i$ summarizes support from path neighbors.
The features $a_i^{in}$ and $a_i^{out}$ describe transition direction, and $z_i$ stores state and schema information.
These groups expand to 12 scalar node features in the released encoder.
They cover semantic score, reciprocal rank, entry and tool frequency, two anchor-transition directions, anchor co-occurrence, two schema-flow directions, state fit, missing-field count, and anchor-family match.

The relation-aware Transformer computes attention with a learned relation bias.
\begin{align}
    H^0 &= \mathrm{Proj}(X^R),\\
    \alpha_{ij}^{\ell,h}
    &=\mathrm{softmax}_j
    \left(
    \frac{(\mathbf{q}_i^{\ell,h})^\top \mathbf{k}_j^{\ell,h}}{\sqrt{d_h}}
    + \beta_{\ell,h}(r_{ij}^R)
    \right),\\
    H^{\ell+1} &= \mathrm{Block}_{\theta}(H^\ell,\alpha^\ell).
\end{align}
The learned bias $\beta_{\ell,h}(r_{ij}^R)$ injects path relation type into attention.

The encoder then predicts membership and path-position scores.
\begin{align}
    \hat y_i^{mem} &= \sigma(w_m^\top h_i^L),\\
    \hat y_{i,r}^{pos} &=
    \frac{\exp(w_r^\top h_i^L)}{\sum_j \exp(w_r^\top h_j^L)}.
\end{align}
Set decoding balances membership, path-slot coverage, missing-field coverage, and semantic diversity.
Let $g_{if}$ indicate that tool $i$ produces missing field $f$, and let $E_i$ be its normalized tool representation.
For a partial set $\candidate$, the decoder adds the remaining tool with the largest marginal score
\begin{align}
    G_i^{slot}
    &=\sum_r\left[\max\{c_r,\hat y_{i,r}^{pos}\}-c_r\right],\\
    G_i^{field}
    &=\frac{1}{\max\{1,|F_q^{miss}|\}}\notag\\
    &\quad\times\sum_{f\in F_q^{miss}}
      \left[\max\{c_f,g_{if}\}-c_f\right],\\
    D_i&=
    \begin{cases}
      0, & \candidate=\varnothing,\\
      \max_{j\in\candidate}E_i^\top E_j, & \text{otherwise},
    \end{cases}\\
    s_i^R&=\hat y_i^{mem}
    +\lambda(G_i^{slot}+G_i^{field})\notag\\
    &\quad-\eta_{\mathrm{div}}D_i,
    \label{eq:set-decoding}
\end{align}
where $c_r$ and $c_f$ are the current maxima in $\candidate$.
The missing-field set collects required fields in the 128-tool frontier that are absent from $F_0(q)$.
The missing-field gain is zero when this set is empty.
The diversity term is zero when $\candidate$ is empty.
The decoder repeats this update until it has selected all 32 tools in the final menu.

\subsection{State-Path Reranker}

The reranker receives the bounded set $\candidate_K$.
It constructs $C_q^O=(V_q^O,R_q^O)$ over exactly these 32 tools.
\begin{equation}
    V_q^O=\candidate_K.
\end{equation}

Each candidate feature is
\begin{equation}
    \begin{aligned}
    x_i^O=
    [&s_i^R,\;\mathrm{rank}_i,\;\hat y_i^{mem},\;\hat y_{i,\cdot}^{pos},\\
     &\chi_i(s_0),\;\mathrm{miss}_i,\;\mathrm{fam}_i],
    \end{aligned}
    \label{eq:reranker-candidate-feature}
\end{equation}
where $s_i^R$, $\mathrm{rank}_i$, and $\hat y_{i,\cdot}^{pos}$ come from the retriever.
This vector contains 14 scalar node features.

Each ordered pair receives the following feature vector.
\begin{equation}
    \begin{aligned}
    a_{ij}^O=
    [&\Delta_{ij}^{rank},\Delta_{ij}^{sim},\rho(i,j),t_{ij},t_{ji},\\
     &\kappa_{ij}^{schema},\kappa_{ji}^{schema},h_{ij}].
    \end{aligned}
    \label{eq:reranker-pair-feature}
\end{equation}
Here $t_{ij}$ indicates an observed forward training transition.
The 9-dimensional vector $h_{ij}$ contains shared service, family, and name indicators, description overlap, three field-overlap counts, and two frequency differences.
Together these terms form the 16-dimensional continuous attention bias.
Field-level availability is carried separately as deterministic decoder context for prefix state fit.

The reranker uses these entries as a continuous pair bias in attention.
\begin{align}
    Z^0 &= \mathrm{Proj}(X^O),\\
    \bar{\mathbf q}_i^{\ell,h} &= \bar{W}_Q^{\ell,h}z_i^\ell,\quad
    \bar{\mathbf k}_j^{\ell,h}=\bar{W}_K^{\ell,h}z_j^\ell,\\
    \alpha_{ij}^{\ell,h}
    &=\mathrm{softmax}_j\!\Bigg(
        \frac{(\bar{\mathbf q}_i^{\ell,h})^\top \bar{\mathbf k}_j^{\ell,h}}{\sqrt{d_h}}
        \nonumber\\[-1mm]
    &\hspace{31mm} +(w_e^{\ell,h})^\top a_{ij}^O
    \Bigg),\\
    Z^{\ell+1} &= \mathrm{Block}_{\psi}(Z^\ell,\alpha^\ell).
\end{align}
The resulting representation feeds the four prediction heads used in the main text.
\begin{align}
    e_i &= \mathrm{Entry}(z_i),\\
    m_i &= \mathrm{Member}(z_i),\\
    \phi_{i,r} &= \mathrm{Slot}(z_i,r),\\
    b_{ij} &= \mathrm{Prec}(z_i,z_j,a_{ij}^O).
\end{align}

The reranker is trained with the following objective.
\begin{equation}
\begin{aligned}
    \mathcal{L}_O
    =&\;\lambda_{entry}\mathcal{L}_{entry}
    +\lambda_{slot}\mathcal{L}_{slot}
    +\lambda_{prec}\mathcal{L}_{prec}\\
    &+\lambda_{member}\mathcal{L}_{member}
    +\lambda_{path}\mathcal{L}_{path}.
\end{aligned}
\label{eq:reranker-loss}
\end{equation}
Together, the five terms supervise entry selection and membership, path slots and precedence, and the executability of the leading prefix.

The final menu maximizes Eq.~\ref{eq:main-rerank-score} by placing the leading 8 tools left to right with development-selected constants.
The leading prefix is the proposed execution path.
Remaining tools retain their retriever order and are appended as backups.

\section{Complete Cross-Benchmark Results}

The main text reports one primary measure from each benchmark.
Table~\ref{tab:full-cross-benchmark} collects the accompanying task and path measures for readers who want the full comparison.

\begin{center}
\centering
\captionof{table}{\textbf{Complete task and path results across five benchmarks.}
Bold and underline mark the best and second-best values within each reported measure.}
\label{tab:full-cross-benchmark}
\scriptsize
\setlength{\tabcolsep}{1.0pt}
\begin{tabular}{@{}L{0.285\linewidth}*{6}{C{0.108\linewidth}}@{}}
\toprule
Menu constructor
& \multicolumn{2}{c}{\toolbench{}}
& \multicolumn{2}{c}{AppWorld}
& \multicolumn{2}{c}{TRAJECT-Bench} \\
\cmidrule(lr){2-3}\cmidrule(lr){4-5}\cmidrule(l){6-7}
& Succ. & Chain
& R@1 & R@5
& Chain & First \\
\midrule
COLT \citep{Qu2024COLT}
& 0.582 & 0.477
& 0.190 & 0.309
& 0.230 & 0.720 \\
ToolRet \citep{Shi2025ToolRet}
& 0.632 & 0.401
& 0.114 & 0.144
& 0.188 & 0.564 \\
Tool-REX family \citep{Lu2026ToolREX}
& 0.661 & 0.533
& 0.180 & 0.299
& 0.350 & 0.771 \\
SkillRouter \citep{Zheng2026SkillRouter}
& 0.707 & 0.510
& 0.123 & 0.279
& 0.167 & 0.545 \\
ToolGen \citep{Wang2025ToolGen}
& 0.730 & 0.510
& 0.049 & 0.243
& 0.708 & 0.838 \\
\midrule
\textbf{State-Path (ours)}
& \textbf{0.898} & 0.704
& \textbf{0.465} & 0.505
& \textbf{0.732} & 0.837 \\
\bottomrule
\end{tabular}

\vspace{4pt}
\begin{tabular}{@{}L{0.34\linewidth}*{4}{C{0.15\linewidth}}@{}}
\toprule
Menu constructor
& \multicolumn{2}{c}{UniToolCall}
& \multicolumn{2}{c}{ToolHop} \\
\cmidrule(lr){2-3}\cmidrule(l){4-5}
& Prefix & First & First & Path \\
\midrule
COLT \citep{Qu2024COLT} & 0.215 & 0.407 & 0.238 & 0.557 \\
ToolRet \citep{Shi2025ToolRet} & 0.134 & 0.279 & 0.196 & 0.538 \\
Tool-REX family \citep{Lu2026ToolREX} & 0.262 & 0.488 & 0.597 & 0.733 \\
SkillRouter \citep{Zheng2026SkillRouter} & 0.169 & 0.343 & 0.467 & 0.662 \\
ToolGen \citep{Wang2025ToolGen} & 0.215 & 0.355 & 0.265 & 0.590 \\
ToolACE \citep{Liu2025ToolACE} & \underline{0.297} & -- & -- & -- \\
\midrule
\textbf{State-Path (ours)} & \textbf{0.635} & 0.744 & \textbf{0.683} & \textbf{0.756} \\
\bottomrule
\end{tabular}
\end{center}

\section{Recent Dependency-Aware Planners}
\label{sec:recent-planner-comparison}

This comparison asks how recent dependency-aware planners perform when they must provide one menu before execution.
We use AutoTool and Dynamic Tool Dependency Retrieval to construct a 32-tool menu \citep{AutoTool2025,Patel2026DynamicToolDependency}.
ToolTree's tree search is evaluated through the same 32-tool interface \citep{ToolTree2026}.
On the 304 \toolbench{} tasks, AutoTool, Dynamic Tool Dependency Retrieval, and ToolTree reach complete-chain coverage of 0.592, 0.388, and 0.609, respectively.
State-Path reaches 0.704 with the same 32-tool interface and task set.

\section{AssistantBench Public-Validation Diagnostic}
\label{sec:assistantbench-validation}

AssistantBench provides an independent web-answer scorer for testing whether state-aware evidence routing helps a shared answer generator \citep{Yoran2024AssistantBench}.
We compare routing policies on the same 33 public validation tasks.
Every policy uses the same web cache, top-10 page budget, answer generator, and official BrowserGym scorer.
Only the routing policy changes across rows.
Table~\ref{tab:assistantbench-strong-baselines} reports the shared scorer and paired outcomes for each policy.

\begin{center}
\begin{minipage}{\columnwidth}
\centering
\captionof{table}{\textbf{Comparison of evidence-routing policies on AssistantBench public validation.}
Pairwise entries give State-Path wins, losses, and ties against each baseline.}
\label{tab:assistantbench-strong-baselines}
\scriptsize
\setlength{\tabcolsep}{2.2pt}
\begin{tabular}{@{}L{0.43\linewidth}C{0.13\linewidth}C{0.13\linewidth}C{0.20\linewidth}@{}}
\toprule
Evidence routing policy & Hit & Acc. & Win/Loss/Tie \\
\midrule
E5 dense retrieval & 45.5 & 17.9 & 17/4/12 \\
BGE reranker & 48.5 & 16.7 & 18/4/11 \\
Qwen3 reranker & 42.4 & 22.7 & 15/6/12 \\
ColBERT late interaction & 48.5 & 17.8 & 16/5/12 \\
HyDE hypothetical retrieval & 48.5 & 18.7 & 16/4/13 \\
RAG-Fusion routing & 48.5 & 19.4 & 16/5/12 \\
Hybrid RRF routing & 45.5 & 15.1 & 18/4/11 \\
RankGPT listwise reranking & 42.4 & 23.4 & 15/3/15 \\
State-Path routing & \textbf{51.5} & \textbf{42.9} & ref. \\
\bottomrule
\end{tabular}
\end{minipage}
\end{center}

State-Path reaches 42.9 accuracy versus 23.4 for RankGPT, with 15 wins, 3 losses, and 15 ties.
The smaller evidence-hit gain suggests that route structure also helps the generator use retrieved pages.

\end{document}